\documentclass[10pt,twocolumn,letterpaper]{article}

\usepackage{cvpr}              
\definecolor{cvprblue}{rgb}{0.21,0.49,0.74}
\usepackage[pagebackref,breaklinks,colorlinks,allcolors=cvprblue]{hyperref}

\usepackage{bm} %

\usepackage{amsmath}
\AtBeginDocument{%
  \setlength{\abovedisplayskip}{6pt}%
  \setlength{\belowdisplayskip}{6pt}%
  \setlength{\abovedisplayshortskip}{4pt}%
  \setlength{\belowdisplayshortskip}{4pt}%
}
\usepackage[utf8]{inputenc}

\usepackage{booktabs}
\usepackage[table]{xcolor}  

\usepackage{enumitem}
\usepackage{amsmath,amssymb} 

\usepackage{multirow}
\usepackage{colortbl}
\usepackage[table]{xcolor}

\usepackage{booktabs}    
\usepackage{makecell}    

\usepackage[capitalize]{cleveref}
\usepackage{placeins}

\def\paperID{****} 
\def\confName{CVPR}
\def\confYear{2026}

\title{ReManNet: A Riemannian Manifold Network for Monocular 3D Lane Detection}

\author{
Chengzhi Hong \quad Bijun Li \{Corresponding author\} \\
State Key Laboratory of Information Engineering in Surveying, Mapping and Remote Sensing \\
Wuhan University, Wuhan, China \\
\texttt{ht005305@whu.edu.cn} \quad \texttt{lee@whu.edu.cn}
}

\begin{document}
\maketitle
\begin{abstract}
Monocular 3D lane detection remains challenging due to depth ambiguity and weak geometric constraints. Mainstream methods rely on depth guidance, BEV projection, and anchor- or curve-based heads with simplified physical assumptions, remapping high-dimensional image features while only weakly encoding road geometry. Lacking an invariant geometric-topological coupling between lanes and the underlying road surface, 2D-to-3D lifting is ill-posed and brittle, often degenerating into concavities, bulges, and twists. To address this, we propose the Road-Manifold Assumption: the road is a smooth 2D manifold in $\mathbb{R}^3$, lanes are embedded 1D submanifolds, and sampled lane points are dense observations, thereby coupling metric and topology across surfaces, curves, and point sets. Building on this, we propose ReManNet, which first produces initial lane predictions with an image backbone and detection heads, then encodes geometry as Riemannian Gaussian descriptors on the symmetric positive-definite (SPD) manifold, and fuses these descriptors with visual features through a lightweight gate to maintain coherent 3D reasoning. We also propose the 3D Tunnel Lane IoU (3D-TLIoU) loss, a joint point-curve objective that computes slice-wise overlap of tubular neighborhoods along each lane to improve shape-level alignment. Extensive experiments on standard benchmarks demonstrate that ReManNet achieves state-of-the-art (SOTA) or competitive results. On OpenLane, it improves F1 by +8.2\% over the baseline and by +1.8\% over the previous best, with scenario-level gains of up to +6.6\%. The code will be publicly available at \url{https://github.com/changehome717/ReManNet}. 
\end{abstract}    
\section{Introduction}
\label{sec:intro}
3D lane detection is a core perception task for autonomous
driving because it provides metric lane geometry and semantics
for downstream planning~\cite{zheng2023path, 10994357}, lane keeping \cite{lanekeeping,10969156}, and scene understanding \cite{zhi2024dynamic,lv2025t2sg}. Although LiDAR offers reliable depth, its high cost and weak responses on road markings limit large-scale deployment. In contrast, monocular cameras are inexpensive and capture rich appearance cues, motivating a shift toward monocular 3D lane detection. Nevertheless, recovering accurate lane geometry from a single image remains highly challenging without explicit depth supervision or principled geometric constraints.

Existing monocular 3D lane detectors can be broadly grouped into
three categories. \textbf{(i) Depth-guided methods}~\cite{once,huang2022monodtrdepth,depth2,lanecppcurve4} infer depth or
voxelized features before lifting image evidence into 3D,
making performance sensitive to depth quality. \textbf{(ii) Bird's-eye-view (BEV)-centric models}~\cite{chen2022persformerbev1, garnett20193dbev2, guo2020genbev3, liu2022learningbev4} regress lanes from image-to-BEV features, but
their implicit planarity assumptions can introduce systematic bias
on nonplanar roads. \textbf{(iii) Line-modeling approaches} parameterize lanes using explicit anchors~\cite{huang2023anchor3dlaneanchor1, huang2024anchor3dlane++anchor2, zheng2024pvalaneanchor3}, polynomial curves~\cite{bai2022curveformercuvr1, kalfaoglu2024topobdacurve2, pittner20233dcurve3, bai2025curveformer++curve5}, or grouped keypoints~\cite{wang2023bevkey1, ozturk2025glane3dbev5key2}. However, under challenging visual conditions, missing or corrupted local cues can lead to erroneous or incomplete point predictions, making the matching between detected lane evidence and the optimized line model unstable and thereby degrading overall accuracy.

Despite their different formulations, existing methods share a common design tendency: they prioritize 2D image features as the primary predictive signal, investing heavily in image-derived intermediate representations such as depth maps, BEV features, or planar reconstructions. By contrast, explicit 3D lane coordinates are relegated to an auxiliary role: as region‑of‑interest (RoI) sampling targets~\cite{huang2023anchor3dlaneanchor1,zheng2024pvalaneanchor3}, as supervision during training, and as weak geometric regularizers (e.g., parallelism~\cite{huang2023anchor3dlaneanchor1,huang2024anchor3dlane++anchor2}, smoothness~\cite{lanecppcurve4}, or projection of 3D structural priors into 2D~\cite{li2022reconstruct}). This practice underutilizes 3D coordinates as carriers of metric constraints and topological structure. Lacking metric and topological invariants, the induced geometric space becomes unstable and 2D‑to‑3D lifting is ill‑conditioned. Learning directly from high‑dimensional observations without explicit structure is brittle and suffers from the curse of dimensionality~\cite{donoho2000highcurse}. Consequently, lifting predicted lanes to 3D can yield structural collapse in the reconstructed road space, manifesting as spurious concavities, bulges, and twists.

To preserve the metric and topological invariants of road space, we ground our formulation in principles of road geometric design~\cite{wolhuter2015geometricroaddesign1}, which enforce alignment continuity and gradual variation of curvature and grade. Consequently, despite global undulations, each local neighborhood of the roadway is well approximated by a smooth, nonsingular surface. Building on this observation, we posit the Road-Manifold Assumption: the roadway is a smooth, embedded two‑dimensional manifold $\mathcal{M}\subset\mathbb{R}^3$ admitting a smooth atlas of Euclidean charts, and lanes are smooth one‑dimensional submanifolds $\gamma\subset\mathcal{M}$ that inherit both local smoothness and global coherence from $\mathcal{M}$. Lane points are treated as sufficiently dense samples on these submanifolds. Equipped with the ambient Euclidean metric $g$ restricted to $\mathcal{M}$ by the pullback~\cite{spd1}, the pair $(\mathcal{M},g)$ forms a Riemannian manifold~\cite{lin2008riemannianmanifoldlearning} that provides intrinsic distances and enables coordinate‑invariant objectives and regularizers. This representation supports principled dimensionality reduction~\cite{law2006incrementalreduction} and facilitates learning and inference in a low-dimensional space while preserving geometric and topological fidelity. To encode intrinsic structural correlations on $\mathcal{M}$, we model local relationships in tangent spaces using symmetric positive definite (SPD) matrices, which are widely used in medical imaging~\cite{medical1, medical2}, signal processing~\cite{signal1, signal2}, and computer vision~\cite{cv1, cv2, cv3}.

Although numerous neural networks on SPD manifolds (denoted by $\mathrm{Sym}_+^{n}$)~\cite{spd1, wang2024spd3, spd4, spd5, spd7, spd8} have been proposed, many apply Riemannian operations without establishing or evaluating that the data admit a well‑defined manifold structure. We make this requirement explicit via the Road‑Manifold Assumption, which provides the geometric foundation of our approach. Building on this premise and leveraging visual cues, we introduce ReManNet, a Riemannian manifold network for monocular 3D lane detection that fuses visual features with geometric representations on $\mathrm{Sym}_+^{n}$. Concretely, the network predicts raw 3D lane points using an image backbone and detection heads, then strengthens spatial relations using a position‑weighted convolution layer. It estimates local SPD covariances~\cite{nguyen2021geomnetspd1action1} to construct Riemannian Gaussian~\cite{pennec2004probabilitiesgsstatistic} descriptors and aligns them via parallel transport~\cite{zhang2013probabilisticpt1} along geodesics induced by the affine-invariant Riemannian metric (AIRM)~\cite{ilea2018covarianceairm}. For numerically stable Euclidean processing, the descriptors are mapped to the Lie algebra via the matrix logarithm, vectorized, and projected to compact fusion features~\cite{pennec2006riemannianlog}. A gating module adaptively fuses the visual and geometric streams. For holistic supervision, we introduce the 3D Tunnel Lane IoU (3D-TLIoU) loss, which measures slice-wise overlap of tubular neighborhoods along each lane, providing shape‑level guidance and improving metric accuracy.

Our contributions are summarized as follows:
\begin{itemize}[leftmargin=*, itemsep=0pt, topsep=0pt]


\item \textbf{Road-Manifold Assumption.} We introduce the Road-Manifold Assumption, which formalizes road space as a smooth 2D manifold in $\mathbb{R}^3$ with lane markings as 1D submanifolds and points as dense samples. This yields a consistent representation of metric and topological structure across road surfaces, lane curves, and sampled points.

\item \textbf{ReManNet.} We propose a Riemannian manifold network that bootstraps image-based coordinate proposals, encodes lane geometry as Riemannian Gaussian descriptors on the $\mathrm{Sym}_+^{n}$, enforces geodesic consistency via AIRM‑based parallel transport, and fuses geometric descriptors with image features through a gating module to produce robust 3D lane predictions.
\item \textbf{3D Tunnel Lane IoU (3D-TLIoU) Loss.} We introduce a slice‑wise IoU between tubular neighborhoods swept along each lane, delivering holistic, shape‑level supervision beyond conventional point‑wise distance losses.
\item \textbf{State-of-the-art performance.}  ReManNet achieves leading results on standard benchmarks. On OpenLane, it improves the F1 score by +8.2\% over the baseline and by +1.8\% over the previous best, while achieving the best category accuracy and localization performance. In scene-level evaluations, it surpasses existing methods in most scenarios, with a maximum F1 gain of +6.6\%.
\end{itemize}

\begin{figure*}[t]
    \centering
    \includegraphics[width=\linewidth]{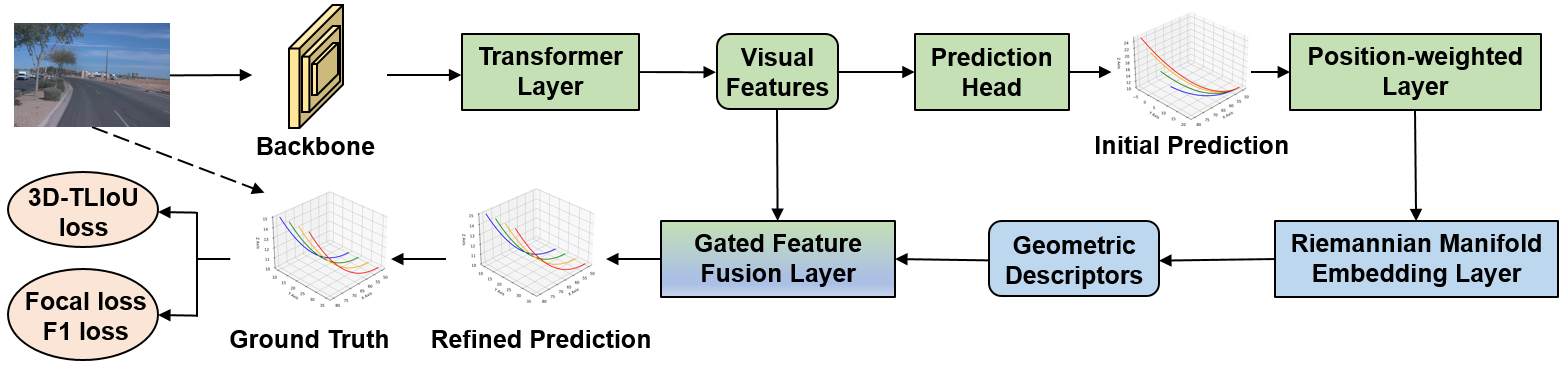}
    \caption{Overall architecture of ReManNet. A backbone and transformer generate initial lane predictions. A position-weighted layer encodes spatial context, and a Riemannian manifold embedding layer maps the resulting features to SPD Gaussian descriptors. A gated visual-geometric fusion layer combines these descriptors with visual features to produce refined predictions. The entire network is trained end-to-end using our geometry-consistent 3D-TLIoU loss alongside standard regression and classification objectives.
}
    \label{fig:spd_network}
\end{figure*}

\section{Related Work}
\label{sec:Related}
 \subsection{3D Lane Detection}
Estimating the 3D geometry of lanes is a fundamental task in autonomous driving, serving as a critical prerequisite for downstream navigation and planning~\cite{Zhang31122024, zheng2023autonomous, ma2025modeling}. A direct route is to acquire depth with specialized sensors ~\cite{sensor1, sensor2}, but hardware cost and the maintenance burden of multi-sensor rigs hinder large-scale deployment. These constraints motivate monocular 3D lane detection, which infers 3D coordinates from a single RGB image.

In this setting, depth-guided regression first predicts a depth field and then recovers the 3D layout~\cite{once, huang2022monodtrdepth, depth2, lanecppcurve4}. For example, SALAD~\cite{once} performs 3D lane detection by jointly learning 2D semantic segmentation and monocular depth, then lifting the segmented lane evidence into 3D. However, because coordinate inference depends on an intermediate depth map, accuracy is fundamentally limited by depth quality, and errors propagate through the pipeline.

Another strand projects image features into BEV space~\cite{chen2022persformerbev1, garnett20193dbev2, guo2020genbev3, liu2022learningbev4}. 3DLaneNet~\cite{garnett20193dbev2} applies inverse perspective mapping (IPM) to transform front‑view (FV) features into BEV and regress BEV lane‑anchor offsets. This pipeline tacitly assumes local planarity. Nonplanar geometry, such as ramps, undulating segments, and superelevated curves, systematically warps the BEV representation, propagating 2D feature distortions into 3D metric error.

Alternatively, some methods model lanes as structured line primitives, either through explicit parameterizations, including anchor templates~\cite{huang2023anchor3dlaneanchor1, zheng2024pvalaneanchor3, huang2024anchor3dlane++anchor2}, polynomial or spline curves~\cite{bai2022curveformercuvr1, kalfaoglu2024topobdacurve2, lanecppcurve4}, and keypoint detection \cite{ozturk2025glane3dbev5key2, wang2023bevkey1} followed by polyline association. Anchor3DLane \cite{huang2023anchor3dlaneanchor1} projects anchors onto the FV feature to extract anchor-aligned features. LaneCPP \cite{lanecppcurve4} employs a B-spline parameterization for detection. GLane3D \cite{ozturk2025glane3dbev5key2} detects lanes as a graph of keypoints. Many methods also introduce geometric priors such as nominal lane width~\cite{ai2022equalwidth}, parallelism~\cite{huang2023anchor3dlaneanchor1}, and surface smoothness~\cite{lanecppcurve4}. These priors
can stabilize optimization, but they do not hold universally under lane additions, drops, variable widths, or sharp curvature
transitions.

Reconstruction-centric \cite{luo2023latr,li2022reconstruct} and graph-optimization \cite{ozturk2025glane3dbev5key2} methods serve as global formulations of geometric priors: they learn geometry-aware embeddings and enforce local connectivity, yet recover surfaces extrinsically in Euclidean space. The resulting geometry inherits a chordal Euclidean metric rather than the intrinsic geodesic metric on the road surface. This metric incompatibility precludes local isometry and curvature preservation, conflates geodesic neighborhoods with their Euclidean counterparts, and induces normal-direction smoothing akin to mean curvature flow. Thin structures consequently collapse, curvature estimates become biased, and spurious topological shortcuts emerge, degrading stability and physical interpretability.

 \subsection{Manifold Learning}
 Manifold learning addresses geometric representation and nonlinear dimensionality reduction by assuming that high‑dimensional observations lie near a low‑dimensional smooth manifold. The goal is to preserve intrinsic geometry such as distances and neighborhood relations, which mitigates the curse of dimensionality~\cite{donoho2000highcurse} and surpasses linear methods like Principal Component Analysis (PCA)~\cite{jolliffe2011principalpca1, turk1991eigenfacespca2}, Multidimensional Scaling (MDS)~\cite{cox2008multidimensionalmds1}, and Linear Discriminant Analysis (LDA)~\cite{duda2006patternlda1}. Representative approaches include Isometric Feature Mapping (ISOMAP)~\cite{tenenbaum2000globalisomap}, which preserves geodesic distances; Locally Linear Embedding (LLE)~\cite{roweis2000nonlinearlle}, which reconstructs each sample from its nearest neighbors; and Laplacian Eigenmaps~\cite{belkin2003laplacian}, which constructs a spectral embedding via the graph Laplacian.

 Correlations in non-Euclidean data are naturally modeled on $\mathrm{Sym}_+^{n}$. Equipped with Riemannian metrics such as AIRM~\cite{ilea2018covarianceairm}, the Log-Euclidean metric (LEM)~\cite{arsigny2005fastlem}, and the Log Cholesky metric (LCM)~\cite{lin2019riemannianlcm}, SPD geometry supplies geodesic distances and stable tangent‑space operations. This supports learning that is globally nonlinear yet locally linear, and has proven effective in a range of tasks such as action recognition~\cite{nguyen2021geomnetspd1action1, zhang2020deepspdaction2}, scene classification~\cite{sun2017learningscenerego1, wang2021symnetscenerego2}, and facial emotion recognition~\cite{brooks2019riemannianface1, spd2}.

  Within this framework, Gaussian embeddings provide principled summaries of local distributions, uncertainty, and correlations. In statistics and computer vision, they are used to compare distributions and build geometry‑aware features. A prevalent strategy identifies Gaussian families with SPD matrices by endowing the covariance space with a Riemannian symmetric‑space structure, yielding closed‑form distances and tractable embeddings~\cite{lovric2000multivariatergincv}. Related work connects Gaussians to subspaces of AIRM~\cite{ilea2018covarianceairm} and leverages Log-Euclidean geometry~\cite{arsigny2005fastlem} for efficient linear embeddings in the log domain~\cite{li2016locallinerembed}. Beyond Euclidean domains, Gaussian models have been generalized to $\mathrm{Sym}_+^{n}$, symmetric spaces~\cite{said2017riemanniansym1}, and homogeneous spaces~\cite{turaga2008statisticalhomo1}.

  These advances motivate the use of Riemannian Gaussians as intrinsic descriptors. They preserve coordinate invariance, capture anisotropy and correlation, and admit principled parallel transport on the manifold. In 3D lane detection, such descriptors summarize local neighborhoods in a way that respects the underlying geometry, enabling robust feature fusion and supporting coherent reasoning.
 \section{Method}
 \subsection{Lane Representation}

The goal of 3D lane detection is to predict a set of 3D lanes from a single input image $\mathbf{I} \in \mathbb{R}^{3 \times H \times W}$. We represent the lane set as \(\mathbf{\Omega} = \{ \mathbf{L}_j \}_{j=1}^{K}\), where \(K\) is the number of lane instances. Each lane \(\mathbf{L}_j = (\mathbf{P}_j, C_j)\) consists of a category label \(C_j\) and a fixed-length point sequence \(\mathbf{P}_j = \{ (x_i^j, y_i, z_i^j) \}_{i=1}^{Q}\), where \(Q\) is the predefined number of sampled points. Following prior work~\cite{chen2022persformerbev1, guo2020genbev3}, the longitudinal coordinates are sampled from a predefined reference set \(Y_{\text{ref}} = \{ y_i \}_{i=1}^{Q}\), which is shared across all lanes.

\subsection{Riemannian Manifold Neural Network}
\subsubsection{Method Overview}
Figure~\ref{fig:spd_network} shows the overall architecture of
ReManNet. 
Given an input image $\mathbf{I} \in \mathbb{R}^{3 \times H \times W}$, we first adopt an anchor-based pipeline~\cite{huang2023anchor3dlaneanchor1} to obtain an initial set of lane predictions
\begin{equation}
\mathbf{\Omega}^{(0)} = \{\mathbf{L}_j^{(0)} \mid j = 1, \dots, K\}, 
\qquad 
\mathbf{L}_j^{(0)} = (\mathbf{P}_j^{(0)},\, C_j^{(0)}),
\label{eq:init-lanes}
\end{equation}
where the initial 3D point sequence of lane $j$ is
\begin{equation}
\mathbf{P}_j^{(0)} = \{(x_i^{\,j},\, y_i,\, z_i^{\,j})\}_{i=1}^{Q}, 
\qquad 
\{y_i\}_{i=1}^{Q} \subset Y_{\text{ref}}.
\label{eq:init-points}
\end{equation}
We stack the initial lane point sequences into a tensor $\mathbf{X}^{\mathrm{in}}\in\mathbb{R}^{Q\times K\times 3}$. Then feed it into a position-weighted convolutional encoder to extract compact geometric features. For the $i$-th sampled position on lane $j$, the position-weighted convolution is defined as
\begin{equation}
\mathbf{x}^{\mathrm{out}}_{i,j}
= \sum_{i' \in \mathcal{E}_i}
\alpha^{(j)}_{i,i'}\, \widetilde{\mathbf{W}}_{r}\, \mathbf{x}^{\mathrm{in}}_{i',j},
\qquad  r = i' - i \in \{-1,0,+1\},
\label{eq:pwconv}
\end{equation}
where $\mathcal{E}_i = \{i-1,\, i,\, i+1\}$ denotes the local neighborhood along lane $j$. The distance-aware weights are
\begin{equation}
\alpha^{(j)}_{i,i'}
= \frac{\exp\!\left(-\frac{1}{\tau}\, \big| y_i^{\,j} - y_{i'}^{\,j} \big| \right)}
{\sum_{k \in \mathcal{E}_i} \exp\!\left(-\frac{1}{\tau}\, \big| y_i^{\,j} - y_{k}^{\,j} \big| \right)},
\qquad \sum_{i' \in \mathcal{E}_i} \alpha^{(j)}_{i,i'} = 1,
\label{eq:pwalpha}
\end{equation}
where $\tau$ is a learnable temperature parameter. The relative-position kernel is
\begin{equation}
\label{eq:pwkernel}
\widetilde{\mathbf{W}}_{r} =
\begin{cases}
\mathbf{W}_{-1}, & \text{if } r = -1,\\[4pt]
\mathbf{W}_{0},  & \text{if } r = 0,\\[4pt]
\mathbf{W}_{+1}, & \text{if } r = +1,
\end{cases}
\end{equation}
with learnable matrices $\{\mathbf{W}_{-1}, \mathbf{W}_{0}, \mathbf{W}_{+1}\}$.



Based on the encoded lane features, we perform k-means clustering and summarize each cluster with a Gaussian distribution. These Gaussians are then embedded into the SPD manifold through a diffeomorphic mapping~\cite{nguyen2021geomnetspd1action1}, where we compute Riemannian statistics, including the manifold mean, tangent-space covariance, and transported local features~\cite{pennec2004probabilitiesgsstatistic,moakher2005differentialitersolver,zhang2013probabilisticpt1,sra2015conicpt2}. The resulting mean-covariance pairs serve as Riemannian Gaussian descriptors for local lane neighborhoods.

To integrate these descriptors into the network, we map the SPD matrices to the Lie algebra using the matrix logarithm~\cite{vemulapalli2014humanlie} and apply a learnable lower-triangular transform to obtain compact Euclidean features. A gated visual-geometric fusion module then combines these manifold features with image features, followed by classification and regression heads that produce the final predictions. 

The following subsections detail the Riemannian-manifold modules built on the position-weighted encoder.

\subsubsection{SPD Manifold Embedding of Gaussians}
Let $\mathbf{X}^{\mathrm{out}}\in\mathbb{R}^{Q\times K\times d}$ denote the output of the position-weighted convolutional layer.
It consists of $Q\times K$ feature vectors in $\mathbb{R}^d$, which we partition into $S$ groups by K-means,
$\mathbf{F}_{s}=\{\mathbf{f}_{s,m}\}_{m=1}^{q_s}$ for $s=1,\dots,S$.
Within each group, we model the features as approximately i.i.d. samples from a Gaussian
$\mathcal{N}(\boldsymbol{\mu}_s,\boldsymbol{\Sigma}_s)$, capturing local feature consistency / local distributional regularity.
The parameters are estimated by
\begin{align}
\boldsymbol{\mu}_{s} &= \frac{1}{q_s}\sum_{m=1}^{q_s}\mathbf{f}_{s,m}, \\
\boldsymbol{\Sigma}_{s} &= \frac{1}{q_s-1}\sum_{m=1}^{q_s}\big(\mathbf{f}_{s,m}-\boldsymbol{\mu}_{s}\big)\big(\mathbf{f}_{s,m}-\boldsymbol{\mu}_{s}\big)^{\!\top}.
\end{align}
Following the Gaussian-to-SPD construction in~\cite{nguyen2021geomnetspd1action1}, we associate each Gaussian
$\mathcal{N}(\boldsymbol{\mu}_s,\boldsymbol{\Sigma}_s)$ with a unit-determinant SPD
matrix
\begin{equation}
\label{eq:gauss-to-spd}
\resizebox{\linewidth}{!}{$
\mathbf{P}_{s}
= \big(\det(\boldsymbol{\Sigma}_{s})\big)^{-\frac{1}{d+\rho}}
\begin{bmatrix}
\boldsymbol{\Sigma}_{s} + \rho\,\boldsymbol{\mu}_{s}\boldsymbol{\mu}_{s}^\top & \boldsymbol{\mu}_{s}^{(\rho)} \\
(\boldsymbol{\mu}_{s}^{(\rho)})^\top & \mathbf{I}_{\rho}
\end{bmatrix},\quad
\boldsymbol{\mu}_{s}^{(\rho)} := \boldsymbol{\mu}_{s}\mathbf{1}_{\rho}^\top,
$}
\end{equation}
where $\rho$ is a tunable embedding dimension, $\mathbf{1}_{\rho}$ is a $\rho$-dimensional all-ones vector, and $\mathbf{I}_{\rho}$ is the $\rho\times \rho$ identity. $\mathbf{P}_s$ has size $(d+\rho)\times(d+\rho)$.

By the Schur complement~\cite{ouellette1981schur},
\begin{equation}
\det\!\begin{bmatrix} 
\boldsymbol{\Sigma}_{s} + \rho\boldsymbol{\mu}_{s}\boldsymbol{\mu}_{s}^\top & \boldsymbol{\mu}_{s}^{(\rho)} \\ 
(\boldsymbol{\mu}_{s}^{(\rho)})^\top & \mathbf{I}_\rho 
\end{bmatrix}
= \det(\boldsymbol{\Sigma}_s).
\end{equation}
Therefore, the scalar factor in Eq.~\eqref{eq:gauss-to-spd} ensures that $\det(\mathbf{P}_s)=1$, removing scale dependence. Moreover, since the corresponding Schur complement equals the covariance matrix $\boldsymbol{\Sigma}_s \succ 0$, the constructed matrix $\mathbf{P}_s$ is positive definite, i.e., $\mathbf{P}_s \in \mathrm{Sym}_{+}^{\,d+\rho}$.

We implement \eqref{eq:gauss-to-spd} as a Gaussian embedding layer
\begin{equation}
\{\mathbf{P}_{s}\}_{s=1}^{S}
= f_{\text{GaussEmbed}}\!\left(\{\mathbf{X}^{\text{out}}_{i,j}\}\right),
\end{equation}

\subsubsection{Gaussian Statistics on Riemannian Manifolds}
\noindent\textbf{Riemannian Gaussian Statistics.}
Let $\{\mathbf{P}_{s}\}_{s=1}^{S}\subset\mathrm{Sym}_{+}^{n}$
denote the SPD embeddings of feature groups.
We aggregate these elements using the Riemannian mean under the
AIRM \cite{ilea2018covarianceairm}.
The geodesic distance $d_R(\cdot,\cdot)$ is
\begin{equation}
d_R(\mathbf{A},\mathbf{B})
= \left\| \log\!\big(\mathbf{A}^{-1/2}\mathbf{B}\,\mathbf{A}^{-1/2}\big) \right\|_F ,
\label{eq:airm-distance}
\end{equation}
For a candidate SPD point $\mathbf{Y}\in \mathrm{Sym}_{+}^{n}$,
the Riemannian mean is defined as
\begin{equation}
\mathbf{P}_{\mathrm{m}}
= \arg\min_{\mathbf{Y}\in \mathrm{Sym}_{+}^{n}}
\frac{1}{S}\sum_{s=1}^{S} d_R\!\big(\mathbf{Y},\mathbf{P}_{s}\big)^2 .
\label{eq:riemannian-mean}
\end{equation}
It can be computed by the standard log--average--exp iteration~\cite{moakher2005differentialitersolver}
\begin{equation}
\label{eq:karcher-iteration}
\mathbf{Y}^{(t+1)}
=
\operatorname{Exp}_{\mathbf{Y}^{(t)}}\!\Big(
\frac{1}{S}\sum_{s=1}^{S}\operatorname{Log}_{\mathbf{Y}^{(t)}}(\mathbf{P}_{s})
\Big),\qquad \\
\mathbf{P}_{\mathrm{m}}=\lim_{t\to\infty}\mathbf{Y}^{(t)},
\end{equation}
where, under AIRM \cite{ilea2018covarianceairm} and \cite{pennec2006riemannianlog}
\begin{align}
\operatorname{Log}_{\mathbf{Y}}(\mathbf{P})
&=
\mathbf{Y}^{1/2}\,
\log\!\big(\mathbf{Y}^{-1/2}\mathbf{P}\,\mathbf{Y}^{-1/2}\big)\,
\mathbf{Y}^{1/2},\\
\operatorname{Exp}_{\mathbf{Y}}(\mathbf{X})
&=
\mathbf{Y}^{1/2}\,
\exp\!\big(\mathbf{Y}^{-1/2}\mathbf{X}\,\mathbf{Y}^{-1/2}\big)\,
\mathbf{Y}^{1/2}.
\end{align}
Let $\mathbf{X}_s=\operatorname{Log}_{\mathbf{P}_{\mathrm{m}}}(\mathbf{P}_{s})\in
T_{\mathbf{P}_{\mathrm{m}}}\mathrm{Sym}_+^{n}$.
Map each symmetric $\mathbf{X}_s$ to a Euclidean vector via the
norm-preserving half-vectorization $\mathrm{svec}(\cdot)$ (off-diagonals scaled by $\sqrt{2}$),
i.e., $\mathbf{v}_s = \mathrm{svec}(\mathbf{X}_s)\in\mathbb{R}^{p}$ with the dimension of the symmetric vectorization $p=n(n+1)/2$.
The empirical covariance \cite{pennec2004probabilitiesgsstatistic} is
\begin{equation}
\mathbf{P}_{\mathrm{c}}
= \frac{1}{S-1}\sum_{s=1}^{S}
\big(\mathbf{v}_s-\bar{\mathbf{v}}\big)\big(\mathbf{v}_s-\bar{\mathbf{v}}\big)^{\!\top},
\qquad \\
\bar{\mathbf{v}} = \frac{1}{S}\sum_{s=1}^{S}\mathbf{v}_s .
\label{eq:tangent-stats}
\end{equation}

In practice, when $S\le p$ or for numerical stability, a small ridge
$\varepsilon\mathbf{I}_{p}$ can be added to ensure $\mathbf{P}_{\mathrm{c}}\in\mathrm{Sym}_+^{p}$ before
downstream operations.

\textbf{Parallel Transport.}
To express tangent features in a unified coordinate system, we transport each tangent vector to an SPD reference
$\mathbf{P}_{\mathrm{ref}}$.
Given $\mathbf{X}_s=\operatorname{Log}_{\mathbf{P}_{\mathrm{m}}}(\mathbf{P}_{s})\in
T_{\mathbf{P}_{\mathrm{m}}}\mathrm{Sym}_+^n$, we apply parallel transport \cite{zhang2013probabilisticpt1, sra2015conicpt2} along the AIRM geodesic from $\mathbf{P}_{\mathrm{m}}$ to $\mathbf{P}_{\mathrm{ref}}$.
Under AIRM, the transport admits the closed form
\begin{equation}
\label{eq:pt-airm-x}
\widetilde{\mathbf{X}}_s
= \Gamma_{\mathbf{P}_{\mathrm{m}}\to\mathbf{P}_{\mathrm{ref}}}(\mathbf{X}_s)
= \mathbf{C}\,\mathbf{X}_s\,\mathbf{C}^\top .
\end{equation}
\begin{equation}
\label{eq:pt-airm-c}
\begin{aligned}
\mathbf{C}
&= \mathbf{P}_{\mathrm{ref}}^{1/2}
\big(\mathbf{P}_{\mathrm{ref}}^{-1/2}\mathbf{P}_{\mathrm{m}}\mathbf{P}_{\mathrm{ref}}^{-1/2}\big)^{-1/2}
\mathbf{P}_{\mathrm{ref}}^{1/2} \\
&= \mathbf{P}_{\mathrm{m}}^{1/2}
\big(\mathbf{P}_{\mathrm{m}}^{-1/2}\mathbf{P}_{\mathrm{ref}}\mathbf{P}_{\mathrm{m}}^{-1/2}\big)^{1/2}
\mathbf{P}_{\mathrm{m}}^{-1/2}.
\end{aligned}
\end{equation}

\textbf{Riemannian Gaussian Statistics Layer.}
Riemannian statistics aggregation is implemented as a network layer
\begin{equation}
\big\{\mathbf{P}_{\mathrm{m}},\,\mathbf{P}_{\mathrm{c}}\big\}
= f_{\mathrm{RiemStats}}\!\left(
\{\mathbf{P}_{s}\}_{s=1}^{S},\,\mathbf{W}_{\mathrm{pt}}
\right),
\label{eq:riemstats}
\end{equation}
where $\mathbf{W}_{\text{pt}}$ parameterizes a learnable SPD reference
$\mathbf{P}_{\mathrm{ref}}(\mathbf{W}_{\text{pt}})$. The covariance is computed from transported tangent features by substituting the following  $\mathbf{v}_{s}$ to Eq.~\eqref{eq:tangent-stats}
\begin{equation}
\label{eq:riem_cov}
\mathbf{v}_{s} = \mathrm{svec}\!\big(
\mathcal{T}_{\mathbf{P}_{\mathrm{m}}\to \mathbf{P}_{\mathrm{ref}}}^{\,\mathbf{W}_{\text{pt}}}
(\operatorname{Log}_{\mathbf{P}_{\mathrm{m}}}\mathbf{P}_{s})
\big),
\end{equation}
where $\mathcal{T}_{\mathbf{P}_{\mathrm{m}}\to \mathbf{P}_{\mathrm{ref}}}^{\,\mathbf{W}_{\text{pt}}}$ denotes the AIRM parallel transport from $T_{\mathbf{P}_{\mathrm{m}}}\mathrm{Sym}_+^n$ to $T_{\mathbf{P}_{\mathrm{ref}}(\mathbf{W}_{\text{pt}})}\mathrm{Sym}_+^n$. This layer thus outputs the mean-covariance pair $(\mathbf{P}_{\mathrm{m}},\mathbf{P}_{\mathrm{c}})$.

\subsubsection{Network Embedding of Riemannian Gaussians}
Given the mean-covariance pair \((\mathbf{P}_{\mathrm{m}},\mathbf{P}_{\mathrm{c}})\in\mathrm{Sym}_+^{n}\times\mathrm{Sym}_+^{p}\) computed from the Riemannian statistics layer in Eq.~\eqref{eq:riemstats}, we next map this descriptor
to a manifold-aware Lie-group parameterization, which preserves
its first- and second-order statistics while enabling stable
Euclidean processing in the network. 

The Riemannian mean channel is expressed in global coordinates through 
$\zeta:\mathrm{Sym}_{+}^{n}\to\mathbb{R}^{p}$, with
\begin{equation}
\label{eq:zeta}
\zeta(\mathbf{P}_{\mathrm{m}})=\operatorname{svec}(\log\mathbf{P}_{\mathrm{m}}),\qquad 
\zeta^{-1}(u)=\exp(\operatorname{smat}(u)),
\end{equation}
where $\mathrm{svec}(\cdot)$ and $\mathrm{smat}(\cdot)$ denote
symmetric vectorization and its inverse, respectively, with output dimension 
$p = n(n+1)/2$.

Writing the covariance as \(\mathbf{P}_{\mathrm{c}} = \mathbf{L}\mathbf{L}^\top\) with \(\mathbf{L} \in \mathrm{LT}_+^p\) (Cholesky factor \cite{lin2019riemannianlcm}), where $\mathrm{LT}_{p}^{+}$ denotes
the set of full-rank lower-triangular matrices with positive
diagonal entries, we adopt a semidirect-product parameterization of the mean-covariance pair~\cite{nguyen2021geomnetspd1action1}:
\begin{equation}
\label{eq:group-law}
\begin{aligned}
(\mathbf{P}^{\mathrm{m}}_1,\mathbf{P}^{\mathrm{c}}_1)\,\star\,(\mathbf{P}^{\mathrm{m}}_2,\mathbf{P}^{\mathrm{c}}_2)
&=
\Big(
\zeta^{-1}\!\big(\mathbf{L}_2^{\top}\zeta(\mathbf{P}^{\mathrm{m}}_1)+\zeta(\mathbf{P}^{\mathrm{m}}_2)\big),\\[-2pt]
&\hspace{2.1em}
(\mathbf{L}_1\mathbf{L}_2)(\mathbf{L}_1\mathbf{L}_2)^{\top}
\Big),
\end{aligned}
\end{equation}
This formulation couples the mean coordinates and covariance
factors in an associative form suitable for subsequent network
embedding.


A concrete matrix realization is given by the block lower-triangular group
\begin{equation}\label{eq:Bp}
\resizebox{.87\linewidth}{!}{$
\mathcal{B}_{p}^{+}=\left\{
\begin{bmatrix}\mathbf{L}&\mathbf{0}\\ \zeta(\mathbf{P}_{\mathrm{m}})^\top&1\end{bmatrix}
\;\middle|\;
\mathbf{L}\in\mathrm{LT}_+^{p},\ \mathbf{P}_{\mathrm{m}}\in\mathrm{Sym}_+^{n}
\right\}
$}
\end{equation}


Accordingly, we define the embedding layer as
\begin{equation}\label{eq:G-embed}
\resizebox{.88\linewidth}{!}{$
\mathbf{G}=
\begin{bmatrix}
\mathbf{L} & \mathbf{0}\\
\zeta(\mathbf{P}_{\mathrm{m}})^\top & 1
\end{bmatrix},
\qquad \\
\{\mathbf{G}\}=f_{\text{GroupEmbed}}\!\big(\{\mathbf{P}_{\mathrm{m}},\mathbf{P}_{\mathrm{c}}\}\big)
$}
\end{equation}
This representation preserves manifold statistics while
providing an algebraically convenient form for associative
composition and stable Euclidean processing.

\textbf{Triangular Mapping and SPD Projection.} To obtain a compact descriptor for network fusion, we further
apply a learnable lower-triangular transformation
\begin{equation}
\label{eq:learnable_lift}
\mathbf{D}=\mathbf{G}\,\mathbf{W}_{\triangle},\qquad 
\mathbf{W}_{\triangle}\in \mathrm{LT}_+^{\,p+1} \ \ (\mathrm{diag}>0).
\end{equation}
where the positivity of the diagonal of $\mathbf{W}_{\triangle}$ ensures
that $\mathbf{D}$ remains full-rank lower-triangular. We then project
it back to the SPD cone by
\begin{equation}
\mathbf{Q} = \mathbf{D}\mathbf{D}^{\top} \in \mathrm{Sym}_{+}^{p+1}.
\label{eq:spd_proj}
\end{equation}
Working in the log domain locally linearizes the AIRM
geometry. We therefore compute
\begin{equation}
\mathbf{E}=\log(\mathbf{Q}), 
\qquad 
\widetilde{\mathbf{H}}=\operatorname{svec}(\mathbf{E})\in\mathbb{R}^{d_g},\ \ \ d_g=\tfrac{(p+1)(p+2)}{2}.
\label{eq:raw_descriptor}
\end{equation}
Since the raw SPD-log descriptor dimension $d_g$ is
determined by the manifold construction, we use a linear
projection to obtain the final fusion descriptor:
\begin{equation}
\mathbf{H} = \phi_h(\widetilde{\mathbf{H}}) = \widetilde{\mathbf{H}} \mathbf{W}_h
\in \mathbb{R}^{d_h},
\qquad
\mathbf{W}_h \in \mathbb{R}^{d_g \times d_h}.
\label{eq:descriptor_projection}
\end{equation}
The overall pipeline is implemented as
\begin{equation}
\{\mathbf{H}\}
=
f_{\mathrm{SPDProj}}
\!\left(
f_{\mathrm{TriMap}}(\mathbf{G}, \mathbf{W}_{\triangle}), \mathbf{W}_h
\right).
\label{eq:spd_pipeline}
\end{equation}

\subsubsection{Gated Visual--Geometric Feature Fusion}

Let
$\mathbf{F}_{\mathrm{anchor}} \in \mathbb{R}^{B \times A \times d_a}$
denote the anchor-level visual features from the initial prediction module,
where $B$ is the batch size, $A$ is the anchor dimension of visual branch,
and $d_a$ is the visual feature dimension. Let
$\mathbf{H} \in \mathbb{R}^{B \times d_h}$ denote the geometric descriptor produced by Eq.~(\ref{eq:spd_pipeline}).
We broadcast $\mathbf{H}$ along
the anchor dimension so that all anchors of the same sample
share the global geometric context:
\begin{equation}
\mathbf{H}_e = \mathrm{expand}(\mathbf{H}) \in \mathbb{R}^{B \times A \times d_h},
\label{eq:expand_h}
\end{equation}

We then concatenate the visual and geometric features along
the channel dimension:
\begin{equation}
\mathbf{Z} = \mathrm{concat}(\mathbf{F}_{\mathrm{anchor}}, \mathbf{H}_e)
\in \mathbb{R}^{B \times A \times (d_a + d_h)}.
\label{eq:fusion_concat}
\end{equation}

A per-anchor scalar gate is predicted from the concatenated
feature:
\begin{equation}
\mathbf{s}_{\mathrm{gate}} = \mathbf{Z}\mathbf{W}_{\mathrm{gate}}
\in \mathbb{R}^{B \times A \times 1},
\qquad
\mathbf{W}_{\mathrm{gate}} \in \mathbb{R}^{(d_a + d_h)\times 1}.
\label{eq:gate_logits}
\end{equation}
To match the channel dimension of the geometric branch, we
first project the visual anchor feature by
\begin{equation}
\mathbf{F}'_{\mathrm{anchor}} = \mathbf{F}_{\mathrm{anchor}} \mathbf{W}_a
\in \mathbb{R}^{B \times A \times d_h},
\qquad
\mathbf{W}_a \in \mathbb{R}^{d_a \times d_h}.
\label{eq:visual_align}
\end{equation}

With
$\mathbf{g} = \sigma(\mathbf{s}_{\mathrm{gate}}) \in \mathbb{R}^{B \times A \times 1}$,
where $\sigma(\cdot)$ is the element-wise sigmoid, the final
fusion is formulated as an anchor-wise gated residual update:
\begin{equation}
\mathbf{F} = \mathbf{F}'_{\mathrm{anchor}} + \mathbf{g} \odot \mathbf{H}_e
\in \mathbb{R}^{B \times A \times d_h},
\label{eq:gated_fusion}
\end{equation}
where $\odot$ denotes element-wise multiplication with
broadcasting along the channel dimension. In this design,
the visual anchor feature serves as the primary branch,
whereas the geometric descriptor provides a gated residual
correction for anchor-wise refinement. The fused feature
$\mathbf{F}$ is then routed to the subsequent classification
and regression heads.

\begin{table*}[!t]
  \centering
  \small
  \setlength{\tabcolsep}{3.2pt}
  \renewcommand{\arraystretch}{1.12}
  \resizebox{\textwidth}{!}{%
  \begin{tabular}{l c c *{4}{c} *{6}{c}}
    \toprule
    & \multicolumn{6}{c}{\textbf{Overall metrics}} & \multicolumn{6}{c}{\textbf{Scenario F1 (\%)}} \\
    \cmidrule(lr){2-7}\cmidrule(l){8-13}
    \thead{Method} &
    \thead{F1\\\textbf{(\%)}~\(\boldsymbol{\uparrow}\)} &
    \thead{Cate Acc\\\textbf{(\%)}~\(\boldsymbol{\uparrow}\)} &
    \thead{\(\mathbf{Ex/N}\)\\\textbf{(m)}~\(\boldsymbol{\downarrow}\)} &
    \thead{\(\mathbf{Ex/F}\)\\\textbf{(m)}~\(\boldsymbol{\downarrow}\)} &
    \thead{\(\mathbf{Ez/N}\)\\\textbf{(m)}~\(\boldsymbol{\downarrow}\)} &
    \thead{\(\mathbf{Ez/F}\)\\\textbf{(m)}~\(\boldsymbol{\downarrow}\)} &
    \thead{Up \&\\Down~\(\boldsymbol{\uparrow}\)} &
    \thead{Curve~\(\boldsymbol{\uparrow}\)} &
    \thead{Extreme\\Weather~\(\boldsymbol{\uparrow}\)} &
    \thead{Night~\(\boldsymbol{\uparrow}\)} &
    \thead{Intersection~\(\boldsymbol{\uparrow}\)} &
    \thead{Merge \&\\Split~\(\boldsymbol{\uparrow}\)} \\
    \midrule
    3D-LaneNet~\cite{garnett20193dbev2} \scriptsize{[CVPR2019]}   & 44.1 & --   & 0.479 & 0.572 & 0.367 & 0.443 & 40.8 & 46.5 & 47.5 & 41.5 & 32.1 & 41.7 \\
    GenLaneNet~\cite{guo2020genbev3} \scriptsize{[ECCV2020]}  & 32.3 & --   & 0.591 & 0.684 & 0.411 & 0.521 & 25.4 & 33.5 & 28.1 & 18.7 & 21.4 & 31.0 \\
    PersFormer~\cite{chen2022persformerbev1} \scriptsize{[ECCV2022]} & 50.5 & 92.3 & 0.485 & 0.553 & 0.364 & 0.431 & 42.4 & 55.6 & 48.6 & 46.6 & 40.0 & 50.7 \\
    CurveFormer~\cite{bai2022curveformercuvr1} \scriptsize{[ICRA2023]} & 50.5 & -- & 0.340 & 0.772 & 0.207 & 0.651 & 45.2 & 56.6 & 49.7 & 49.1 & 42.9 & 45.4 \\
    LATR~\cite{luo2023latr} \scriptsize{[ICCV2023]} & 61.9 & 92.0 & 0.219 & 0.259 & 0.075 & 0.104 & 55.2 & 68.2 & 57.1 & 55.4 & 52.3 & 61.5 \\
    Anchor3DLane (R18)~\cite{huang2023anchor3dlaneanchor1} \scriptsize{[CVPR2023]}  & 53.7 & 90.9 & 0.276 & 0.311 & 0.107 & 0.138 & 46.7 & 57.2 & 52.5 & 47.8 & 45.4 & 51.2 \\
    Anchor3DLane (R50)$^\dagger$~\cite{huang2023anchor3dlaneanchor1} \scriptsize{[CVPR2023]}   & 57.5 & 91.6 & 0.233 & 0.246 & 0.080 & 0.106 & 52.7 & 60.8 & 56.2 & 54.7 & 49.8 & 56.0 \\
    LaneCPP~\cite{lanecppcurve4} \scriptsize{[CVPR2024]}  & 60.3 & --   & 0.264 & 0.310 & 0.077 & 0.117 & 53.6 & 64.4 & 56.7 & 54.9 & 52.0 & 58.7 \\
    PVALane (R50)~\cite{zheng2024pvalaneanchor3} \scriptsize{[AAAI2024]} & 62.7 & --   & 0.232 & 0.259 & 0.092 & 0.118 & 54.1 & 67.3 & 62.0 & 57.2 & 53.4 & 60.0 \\
    Anchor3DLane++ (R18)~\cite{huang2024anchor3dlane++anchor2} \scriptsize{[TPAMI2024]} & 57.9 & 91.4 & 0.232 & 0.265 & 0.076 & 0.102 & 48.4 & 64.0 & 54.8 & 52.6 & 48.5 & 56.1 \\
    Anchor3DLane++ (R50)~\cite{huang2024anchor3dlane++anchor2} \scriptsize{[TPAMI2024]} & 62.4 & 93.4 & 0.202 & 0.237 & 0.073 & 0.100 & 54.1 & 68.4 & 58.3 & 55.4 & 53.1 & 61.1 \\
    MapTRv2 (R50)~\cite{liao2025maptrv2} \scriptsize{[IJCV2025]} & 53.6 & 88.9 & 0.267 & 0.312 & 0.074 & 0.105 & 50.0 & 53.2 & 54.9 & 51.3 & 43.1 & 53.1 \\
    Glane3D (R18)~\cite{ozturk2025glane3dbev5key2} \scriptsize{[CVPR2025]} & 61.5 & -- & 0.221 & 0.252 & 0.073 & 0.101 & 55.6 & \textbf{69.1} & 56.6 & 56.6 & 52.9 & 61.3 \\
    Glane3D (R50)~\cite{ozturk2025glane3dbev5key2} \scriptsize{[CVPR2025]} & 63.9 & -- & 0.193 & 0.234 & 0.065 & 0.090 & 58.2 & 71.1 & 60.1 & 60.2 & 55.0 & \textbf{64.8} \\ 
    \rowcolor{gray!12}
    ReManNet (R18) (Ours)          & 63.5 & 92.8 & 0.222 & 0.265 & 0.069 & 0.089 & 61.0 & 66.1 & 66.7 & 63.0 & 56.9 & 56.8 \\
    \rowcolor{gray!12}
    ReManNet (R50) (Ours)          & \textbf{65.7} & \textbf{94.7} & \textbf{0.189} & \textbf{0.205} & \textbf{0.060} & \textbf{0.072} & \textbf{63.2} & 67.9 & \textbf{68.6} & \textbf{65.3} & \textbf{60.2} & 59.9 \\
    \bottomrule
  \end{tabular}
  }
\caption{Comparison with SOTA methods on the OpenLane. 
The left block reports overall metrics: F1, category accuracy (Cate Acc), and mean $x$/$z$ errors for the near (N) and far (F) ranges, denoted as $\mathbf{Ex/N}$, $\mathbf{Ex/F}$, $\mathbf{Ez/N}$, and $\mathbf{Ez/F}$. 
The right block reports scenario-wise F1. 
R18 and R50 denote ResNet-18 and ResNet-50 backbones; $^\dagger$ indicates the baseline. 
The best results are shown in \textbf{bold}.}
\label{tab:openlane_combined_noall}
\end{table*}

\begin{table*}[t]
\centering
\large
\setlength{\tabcolsep}{3.0pt}
\renewcommand{\arraystretch}{1.12}
\resizebox{\textwidth}{!}{%
\begin{tabular}{l|c|cccc|c|cccc|c|cccc}
\toprule
\multirow{2}{*}{\makecell{\textbf{Method}}} &
\multicolumn{5}{c|}{\textbf{Balanced Scenes}} &
\multicolumn{5}{c|}{\textbf{Rare Subset}} &
\multicolumn{5}{c}{\textbf{Visual Variations}} \\
\cmidrule(lr){2-6}\cmidrule(lr){7-11}\cmidrule(l){12-16}
& \textbf{F1 (\%)}$\uparrow$ &
\textbf{$\mathbf{Ex/N}$}$\downarrow$ & \textbf{$\mathbf{Ex/F}$}$\downarrow$ & \textbf{$\mathbf{Ez/N}$}$\downarrow$ & \textbf{$\mathbf{Ez/F}$}$\downarrow$ &
\textbf{F1 (\%)}$\uparrow$ &
\textbf{$\mathbf{Ex/N}$}$\downarrow$ & \textbf{$\mathbf{Ex/F}$}$\downarrow$ & \textbf{$\mathbf{Ez/N}$}$\downarrow$ & \textbf{$\mathbf{Ez/F}$}$\downarrow$ &
\textbf{F1 (\%)}$\uparrow$ &
\textbf{$\mathbf{Ex/N}$}$\downarrow$ & \textbf{$\mathbf{Ex/F}$}$\downarrow$ & \textbf{$\mathbf{Ez/N}$}$\downarrow$ & \textbf{$\mathbf{Ez/F}$}$\downarrow$ \\
\midrule
3DLaneNet\,\cite{garnett20193dbev2} \normalsize {[CVPR2019]}    & 86.4 & 0.068 & 0.477 & 0.015 & 0.202 & 72.0 & 0.166 & 0.855 & 0.039 & \textbf{0.521 }& 72.5 & 0.115 & 0.601 & 0.032 & 0.230 \\
GenLaneNet \,\cite{guo2020genbev3} \normalsize {[ECCV2020]}   & 88.1 & 0.061 & 0.496 & 0.012 & 0.214 & 78.0 & 0.139 & 0.903 & 0.030 & 0.539 & 85.3 & 0.074 & 0.538 & 0.015 & 0.232 \\
CLGo \,\cite{liu2022learningbev4}  \normalsize {[AAAI2022]}  & 91.9 & 0.061 & 0.361 & 0.029 & 0.250 & 86.1 & 0.147 & 0.735 & 0.071 & 0.609 & 87.3 & 0.084 & 0.464 & 0.045 & 0.312 \\
PersFormer\,\cite{chen2022persformerbev1}  \normalsize {[ECCV2022]}  & 92.9 & 0.054 & 0.356 & 0.010 & 0.234 & 87.5 & 0.107 & 0.782 & 0.024 & 0.602 & 89.6 & 0.074 & 0.430 & 0.015 & 0.266 \\
GP\,\cite{li2022reconstruct} \normalsize {[CVPR2022]}   & 91.9 & 0.049 & 0.387 & 0.008 & 0.213 & 83.7 & 0.126 & 0.903 & 0.023 & 0.625 & 89.9 & 0.060 & 0.446 & \textbf{0.011} & 0.235 \\
CurveFormer\,\cite{bai2022curveformercuvr1} \normalsize {[ICRA2023]}  & 95.8 & 0.079 & 0.326 & 0.018 & 0.219 & 95.6 & 0.182 & 0.737 & 0.039 & 0.561 & 90.8 & 0.125 & 0.410 & 0.028 & 0.254 \\
LATR\,\cite{luo2023latr}  \normalsize {[ICCV2023]}       & 96.8 & 0.022 & 0.253 & \textbf{0.007} & \textbf{0.202} & 96.1 & 0.050 & 0.600 & \textbf{0.015} & 0.532 & 95.1 & 0.045 & 0.315 & 0.016 & 0.228 \\
Anchor3DLane$^\dagger$\,\cite{huang2023anchor3dlaneanchor1} \normalsize {[CVPR2023]}   & 95.4 & 0.045 & 0.300 & 0.016 & 0.223 & 94.4 & 0.082 & 0.699 & 0.030 & 0.580 & 91.8 & 0.047 & 0.327 & 0.019 & 0.219 \\
Anchor3DLane++ (R18)\,\cite{huang2024anchor3dlane++anchor2}  \normalsize {[TPAMI2024]}  & 96.3 & 0.027 & 0.268 & 0.011 & 0.215 & 96.4 & 0.050 & 0.617 & 0.019 & 0.551 & 92.7 & 0.045 & 0.371 & 0.019 & 0.250 \\
Anchor3DLane++ (R50)\,\cite{huang2024anchor3dlane++anchor2} \normalsize {[TPAMI2024]}   & 96.5 & 0.022 & 0.234 & 0.009 & 0.204 & 96.4 & \textbf{0.043} & 0.580 & 0.017 & 0.529 & 95.3 & 0.035 & 0.292 & 0.012 & 0.229 \\


Glane3D \,\cite{ozturk2025glane3dbev5key2} \normalsize {[CVPR2025]}   & \textbf{98.1} & \textbf{0.021} & 0.250 & 0.007 & 0.213& \textbf{98.4} & 0.044 & 0.621 & 0.023 & 0.566 & 92.7 & 0.046 & 0.364 & 0.020 & 0.317 \\

\rowcolor{gray!12}
ReManNet (R18) (Ours)     & 96.9 & 0.033 & 0.259 & 0.014 & 0.213  & 97.1 & 0.055 & 0.622 & 0.019 & 0.544 & 96.8 & \textbf{0.033} & 0.272 & 0.013 & 0.225 \\
\rowcolor{gray!12}
ReManNet (R50) (Ours)     & 96.9 & 0.035 & \textbf{0.228} & 0.020 & \underline{0.205} & 97.2 & 0.055 & \textbf{0.559} & 0.030 & \textbf{0.521} & \textbf{96.9} & 0.034 & \textbf{0.237} & \underline{0.019} & \textbf{0.213} \\
\bottomrule
\end{tabular}%
}
\caption{Comparison with SOTA methods on the ApolloSim. Results are reported on the \textbf{Balanced}, \textbf{Rare}, and \textbf{Visual Variations} subsets. $\mathbf{Ex/N}$, $\mathbf{Ex/F}$, $\mathbf{Ez/N}$, and $\mathbf{Ez/F}$ denote mean $x$/$z$ errors in the near (N) and far (F) ranges.
R18 and R50 denote ResNet-18 and ResNet-50 backbones; $^\dagger$ indicates the baseline. 
The best and competitive results are highlighted in \textbf{bold} and \underline{underlined}, respectively.} 
\label{tab:3d_lane_balanced_rare_variations}
\end{table*}

\subsection{3D Tunnel Lane IoU Loss}
\noindent\textbf{Motivation.}
Conventional point-wise distance losses score errors independently at sampled locations and thus underweight each sample's contribution to global lane geometry. 
This makes optimization sensitive to local outliers and jitter, and encourages overfitting to point annotations.
Motivated by the LineIoU loss \cite{zheng2022clrnet} for 2D lanes, we introduce the 3D Tunnel Lane IoU (3D-TLIoU) loss, which measures the consistency between predicted and ground-truth tubular neighborhoods along the entire lane, thereby promoting shape-level alignment.

\textbf{Definition.}
Sample each lane at $Q$ slices along $Y_{\text{ref}}$.
Let $p_i=(x_i^p,z_i^p)$ and $g_i=(x_i^g,z_i^g)$ denote the predicted and ground-truth points in the slice plane $\{Y=Y_i\}$, and set $d_i=\|p_i-g_i\|$.
Given a tube radius $r_{\text{tube}}>0$, we adopt a simple, monotone slice surrogate for the overlap between equal-radius disks
\begin{equation}
\widehat{\mathrm{IoU}}_i \;=\; \frac{d_{\text{over}}}{\,d_{\text{union}}\,}\;=\;\frac{2r_{\text{tube}}-d_i}{\,2r_{\text{tube}}+d_i\,},
\label{eq:slice}
\end{equation}
 where $\widehat{\mathrm{IoU}}_i$ can be negative to indicate the extent of separation between the two disks. To encode geometric coherence along the lane, define tangents 
$\Delta p_i=p_i-p_{i-1}$ and $\Delta g_i=g_i-g_{i-1}$ $(i\ge2)$, the cosine similarity 
$\mathrm{Sim}_i=\tfrac{\Delta p_i^\top\Delta g_i}{\|\Delta p_i\|\,\|\Delta g_i\|}$,
and the associated penalty $\mathcal{C}_i=\tfrac{1-\mathrm{Sim}_i}{2}\in[0,1]$.

Aggregating slice terms yields a curve-level objective that balances positional proximity and directional agreement. The 3D-TLIoU loss is defined as
\begin{equation}
\mathcal{L}_{\text{3D-TLIoU}}
= 1 -
\frac{\sum_{i=1}^{Q}\!\big(2r_{\text{tube}} - d_i\big)}
     {\sum_{i=1}^{Q}\!\big(2r_{\text{tube}} + d_i\big)}
\;+\;
\lambda_{\mathrm{sim}}\;\frac{1}{Q-1}\sum_{i=2}^{Q}\mathcal{C}_i .
\label{eq:loss}
\end{equation}
This formulation explicitly rewards overlap of tubular neighborhoods while regularizing tangent consistency, yielding robustness to small noise and improved geometric fidelity along the lane.

\textbf{Total loss.}
The training objective combines classification, coordinate regression along $x$ and $z$, and the proposed 3D-TLIoU term
\begin{equation}
\mathcal{L}_{\text{total}}
= \lambda_{\mathrm{cls}}\,\mathcal{L}_{\mathrm{cls}}
\;+\;
\lambda_{\mathrm{reg}}\big(\mathcal{L}_{x}+\mathcal{L}_{z}\big)
\;+\;
\lambda_{\mathrm{tliou}}\,\mathcal{L}_{\mathrm{3D\text{-}TLIoU}} .
\label{eq:total_loss}
\end{equation}
Here, $\lambda_{\mathrm{cls}}$, $\lambda_{\mathrm{reg}}$, and $\lambda_{\mathrm{tliou}}$ are the weights of the classification, regression, and 3D-TLIoU losses, respectively.
\section{Experiment}
\subsection{Datasets and Evaluation Metrics}
\noindent\textbf{Datasets.} OpenLane~\cite{chen2022persformerbev1} is a large-scale real-world benchmark for 3D lane detection built upon the Waymo Open Dataset. It contains 200K frames from 1,000 sequences at a resolution of $1280\times1920$, with over 880K lane annotations covering diverse weather, terrain, and lighting conditions. The benchmark provides per-frame camera intrinsics and extrinsics, and its annotations are comprehensive, including opposite-direction lanes without median barriers. In addition, OpenLane includes category and scene tags, and supports evaluation on a wide range of challenging scenarios, such as uphill and downhill roads, curves, extreme weather, nighttime driving, intersections, and merge/split cases.

ApolloSim \cite{guo2020genbev3} is a photo-realistic synthetic benchmark built in the Unity 3D game engine. It contains 10{,}500 images spanning diverse virtual scenes (highway, urban, residential/downtown, rural) with variability in illumination, weather, traffic/obstacles, and road-surface quality. The dataset is organized into three subsets: Standard, Rare Scenes, and Visual Variations. It also provides per-frame camera parameters for 3D lane evaluation.

\textbf{Evaluation Metrics.}
We follow the evaluation method of~\cite{huang2023anchor3dlaneanchor1}: lanes are uniformly sampled along the longitudinal axis \(Y\) from \(0\) to \(100\,\mathrm{m}\). A predicted lane and a ground-truth lane are matched using a distance threshold $\tau$ (we use $1.5$\,m here); a prediction is counted as a true positive if at least \(75\%\) of its sampled points lie within \(\tau\) of the matched ground truth. 
We report the maximum F1 score under these criteria, together with mean lateral (\(x\)) and vertical (\(z\)) errors computed separately in the near (\(0\)--\(40\,\mathrm{m}\)) and far (\(40\)--\(100\,\mathrm{m}\)) ranges to assess geometric accuracy. On OpenLane, we also report category accuracy, defined as the proportion of true positives with correctly predicted categories.


\subsection{Experimental Settings}
We adopt Anchor3DLane~\cite{huang2023anchor3dlaneanchor1} as our baseline and follow its anchor configuration. The number of sampled points per anchor is set to \(Q = 20\) on OpenLane and \(Q = 10\) on ApolloSim. All input images are resized to \(360 \times 480\), and the backbone outputs feature maps of size \(45 \times 60 \times 64\).

For module hyperparameters, the position-weighted convolutional layer produces \(d = 3\) channels. We apply \(k\)-means clustering with \(S = 30\) clusters and \(20\) iterations. The embedding dimension $\rho$ in Eq.~\eqref{eq:gauss-to-spd} is fixed to 1, and the geometric descriptor \(\mathbf{H}\) has dimension \(d_h = 256\).

For loss functions, the 3D-TLIoU tube radius is set to \(r_{\text{tube}} = 1.5\,\text{m}\) with cosine-similarity weight \(\lambda_{\text{sim}} = 0.4\). The global loss weights are \(\lambda_{\text{cls}} = 1\), \(\lambda_{\text{reg}} = 1\), and \(\lambda_{\text{tliou}} = 0.5\). We train the model on OpenLane and ApolloSim for 60K and 50K iterations, respectively, using the AdanW optimizer (learning rate \(2 \times 10^{-4}\), weight decay \(1 \times 10^{-4}\)), with a batch size of \(4\) and \(8\) data-loader workers. All experiments are conducted on a single NVIDIA GeForce RTX~4090 GPU with an AMD EPYC~9654 CPU.

\subsection{Main Results}
\textbf{OpenLane.} 
Table~\ref{tab:openlane_combined_noall} presents results on the OpenLane dataset. ReManNet with a ResNet-50 backbone achieves SOTA performance, improving F1 by \textbf{+8.2\%} over the Anchor3DLane ResNet-50 baseline and by \textbf{+1.8\%} over the previous best. 
It also achieves the highest category accuracy and the lowest localization errors along \(x\) and \(z\) in both near and far ranges. 
At the scenario level, ReManNet ranks first in most settings, with gains of \textbf{+6.6\%} in \emph{Extreme Weather}, \textbf{+5.2\%} in \emph{Intersection}, \textbf{+5.1\%} in \emph{Night}, and \textbf{+5.0\%} in \emph{Up \& Down}. 
These scenes exhibit weak visual cues or strong geometric variability, and the gains highlight the effectiveness of coupling road-manifold geometric consistency with visual features for 3D reasoning. 
Performance on \emph{Merge \& Split} is lower, 
which may stem from complex topological interactions that 
locally violate manifold consistency. 

\textbf{ApolloSim.} In Table~\ref{tab:3d_lane_balanced_rare_variations}, we report results on the ApolloSim benchmark. ReManNet (R50) delivers the most balanced overall localization performance across the three subsets, particularly in terms of far-range errors. It achieves the lowest far-range $x$-axis error ($\mathrm{Ex/F}$) on all three subsets. Along the $z$ axis, it attains the best far-range error ($\mathrm{Ez/F}$) on \emph{Visual Variations} and a tied-best result on \emph{Rare}, while remaining competitive on \emph{Balanced}. On the \emph{Visual Variations} subset, ReManNet (R50) also achieves the best F1 score, improving by \textbf{+1.6\%} over the previous best, together with clear reductions in far-range errors on both $x$ and $z$. These results suggest that the proposed manifold-based geometric encoding strengthens spatial coherence and robustness under diverse appearance and illumination changes.
\begin{table}[t]
  \centering
  \footnotesize  
  \setlength{\tabcolsep}{4.5pt}  
  \renewcommand{\arraystretch}{1.15}
  \begin{tabular}{c c c | c}
    \toprule
    \textbf{Baseline} & \textbf{3D-TLIoU Loss} & \textbf{Riemannian Gaussian} & \textbf{F1 (\%)} $\uparrow$ \\
    \midrule
    \checkmark &  &  & 57.5 \\
    \checkmark & \checkmark &  & 60.5 \\
    \checkmark &  & \checkmark & 62.0 \\
    \checkmark & \checkmark & \checkmark & \textbf{65.7} \\
    \bottomrule
  \end{tabular}
  \caption{Ablation study of ReManNet components.}
  \label{tab:ablation_Components}
\end{table}

\begin{table}[t]
  \centering
  \footnotesize  
  \setlength{\tabcolsep}{4.5pt}  
  \renewcommand{\arraystretch}{1.15}
  \begin{tabular}{c c c | c}
    \toprule
    \textbf{Baseline} & \textbf{$\mathcal{C}_i$} & \textbf{3D LIoU term} & \textbf{F1 (\%)} $\uparrow$ \\
    \midrule
    \checkmark &  &  & 57.5 \\
    \checkmark & \checkmark &  & 58.6 \\
    \checkmark &  & \checkmark & 59.9 \\
    \checkmark & \checkmark & \checkmark & \textbf{60.5} \\
    \bottomrule
  \end{tabular}
  \caption{Ablation study of the 3D-TLIoU loss.}
  \label{tab:ablation_loss}
\end{table}
\textbf{Ablation Study.}
To validate the effectiveness of ReManNet components, we conduct four ablation settings on OpenLane: the baseline, adding only the 3D-TLIoU loss, adding only the Riemannian Gaussian module, and the full ReManNet model (Table~\ref{tab:ablation_Components}). 
Introducing the 3D-TLIoU loss yields a \textbf{+3.0\%} F1 improvement by enforcing consistent overlap supervision along lane curves. 
Adding the Riemannian Gaussian module improves F1 by \textbf{+4.5\%}, indicating that encoding geometric descriptors on the SPD manifold enhances spatial reasoning stability. 
Combining both leads to a substantial \textbf{+8.2\%} gain, surpassing the individual contributions and confirming their complementary synergy. 
The joint design allows ReManNet to capture intrinsic geometric consistency and maintain robust 3D reasoning 
under diverse visual conditions, leading to the strongest performance among the ablated variants

For the 3D-TLIoU loss, we ablate its two components, namely the cosine-similarity penalty $\mathcal{C}_i$ and the 3D LIoU term computed between predictions and ground truth. As shown in Table~\ref{tab:ablation_loss}, adding $\mathcal{C}_i$ raises F1 from 57.5\% to 58.6\% (\textbf{+1.1\%}), while using only 3D LIoU yields 59.9\% (\textbf{+2.4\%}). Combining both terms (full 3D-TLIoU Loss) attains 60.5\% (\textbf{+3.0\%} over baseline), outperforming the $\mathcal{C}_i$-only and LIoU-only variants by +1.9\% and +0.6\%. These gains indicate that $\mathcal{C}_i$ (directional alignment) and 3D LIoU (geometric overlap) are complementary and mutually reinforcing, yielding consistent improvements when combined.

\section{Conclusion}
In this work, we introduce the Road-Manifold Assumption to address geometric collapse when reconstructing road surfaces from lane predictions, a problem caused by the lack of intrinsic metric invariance in Euclidean space. Based on this assumption, we build an intrinsically consistent representation that aligns metric and topological structure across road surface, lane curves, and sampled points. We propose ReManNet, a Riemannian manifold network that fuses Riemannian Gaussian descriptors with visual features, alongside a geometry-consistent 3D-TLIoU loss for supervision. Experiments on standard benchmarks demonstrate robust and accurate 3D lane detection with consistent gains over prior methods. Beyond lane detection, we hope this formulation and supervision strategy will inspire geometry‑aware 3D perception, spatial reconstruction, and scene generation.

\section{Acknowledgment}
This work was supported by the Department of Science and Technology of Hubei Province under Grant 2023BAB146.

{
    \small
    \bibliographystyle{ieeenat_fullname}
    \bibliography{main}
}


\end{document}